\documentclass[runningheads]{llncs}
\usepackage{amsmath,bm}
\usepackage{amsfonts}

\def\eqref#1{equation~\ref{#1}}

\def\1{\bm{1}}

\def\ve{{\bm{e}}}

\def\vh{{\bm{h}}}

\def\vp{{\bm{p}}}

\DeclareMathAlphabet{\mathsfit}{\encodingdefault}{\sfdefault}{m}{sl}
\SetMathAlphabet{\mathsfit}{bold}{\encodingdefault}{\sfdefault}{bx}{n}

\newcommand{\R}{\mathbb{R}}

\usepackage{graphicx}
\usepackage{times}
\usepackage{latexsym}
\usepackage[T1]{fontenc}
\usepackage[utf8]{inputenc}
\usepackage{microtype}
\usepackage{svg}
\usepackage{amsmath}
\usepackage{amssymb}
\usepackage{xspace}
\usepackage{tabularx}
\usepackage{subcaption}
\usepackage{algorithm}
\usepackage{algorithmic}
\usepackage{array}
\usepackage{bbding}
\newcommand{\ttbf}[1]{\texttt{\textbf{#1}}}

\newcommand{\toke}[0]{\ttbf{PatReFormer}\xspace}
\newcommand{\datafbn}{FB15K-237}
\newcommand{\datawnn}{WN18RR}
\newcommand{\datadb}{DB100K}
\newcommand{\datayg}{YAGO37}
\newcommand{\method}{\toke}

\usepackage{graphicx}
\graphicspath{ {./images/} }
\usepackage{booktabs}

\begin{document}
\title{Separate-and-Aggregate: A Transformer-based Patch Refinement Model for Knowledge Graph Completion}
%
%

\author{Chen Chen\inst{1} \and
Yufei Wang\inst{2} \and Yang Zhang\inst{2} \and Quan Z. Sheng\inst{2} \and
Kwok-Yan Lam\inst{1(}\Envelope\inst{)}}
%
%
\institute{Nanyang Technological University, Singapore \\\email{\{s190009,kwokyan.lam\}@ntu.edu.sg} \and
Macquarie University, Sydney, Australia \\
\email{yufei.wang@students.mq.edu.au} \\
\email{yang.zhang21@hdr.mq.edu.au} \\
\email{michael.sheng@mq.edu.au}}

\titlerunning{A Transformer-based Patch Refinement Model for Knowledge Graph Completion}
\maketitle              
\begin{abstract}

Knowledge graph completion (KGC) is the task of inferencing missing facts from any given knowledge graphs (KG). Previous KGC methods typically represent knowledge graph entities and relations as trainable continuous embeddings and fuse the embeddings of the entity $h$ (or $t$) and relation $r$ into hidden representations of query $(h, r, ?)$ (or $(?, r, t$)) to approximate the missing entities. To achieve this, they either use shallow linear transformations or deep convolutional modules. However, the linear transformations suffer from the expressiveness issue while the deep convolutional modules introduce unnecessary inductive bias, which could potentially degrade the model performance. Thus, we propose a novel Transformer-based Patch Refinement Model (\method) for KGC. \method first segments the embedding into a sequence of patches and then employs cross-attention modules to allow bi-directional embedding feature interaction between the entities and relations, leading to a better understanding of the underlying KG.
We conduct experiments on four popular KGC benchmarks, WN18RR, FB15k-237, YAGO37 and DB100K. The experimental results show significant performance improvement from existing KGC methods on standard KGC evaluation metrics, e.g., MRR and H@n. Our analysis first verifies the effectiveness of our model design choices in \method. We then find that \method can better capture KG information from a large relation embedding dimension. Finally, we demonstrate that the strength of \method is at complex relation types, compared to other KGC models~\footnote{Source code is at~\url{https://github.com/chenchens190009/PatReFormer}}.

\end{abstract}
\section{Introduction}
Knowledge graphs (KGs) have emerged as a powerful tool for representing structured knowledge in a wide range of applications, including information retrieval, question answering and recommendation systems. A typical KG is represented as a large collection of triples $(head\ entity, relation, tail\ entity)$, denoted as $(h, r, t)$. 
Despite having large amount of KG triples, many real-world KGs still suffer from incompleteness issue
. To alleviate this issue, the task of knowledge graph completion (KGC) is proposed~\cite{TransE,RotatE,ConvE,StAR}, which is to predict the missing entity given the query $(h, r, ?)$ or $(?, r, t)$.

Existing methods for KGC generally learn continuous embeddings for entities and relations, with the goal of capturing the inherent structure and semantics of the knowledge graph. They define various scoring functions to aggregate the embeddings of the entity and relation, forming a hidden representation of query $(h, r, ?)$ (or $(?, r, t)$) and determine the plausibility between the query representation and missing entity embedding. Essentially, these scoring functions are a set of computation operations on interactive features of the head entity, relation and tail entity. Early KGC models like TransE~\cite{TransE}, DistMult~\cite{DistMult} and ComplEx~\cite{ComplEx} use simple linear operations, such as addition, subtraction and multiplication. Despite the computational efficiency, these simple and shallow architectures are incapable of capturing complicated features, e.g., poor expressiveness. To improve the model expressiveness, some recent KGC models integrate the deep neural operations into the scoring function. ConvE~\cite{ConvE}, as the start of this trend, applies standard convolutional filters over reshaped embeddings of input entities and relations, and subsequent models~\cite{ConvR,InteractE} follow this trend to further improve the expressiveness of the feature interaction between entities and relations. Although these convolution-based KGC models have achieved significant empirical success, they impose unnecessary image-specific inductive bias (i.e., locality and translation equivariance) to the KGC embedding models, potentially degrading the model performance. 

To combat these limitations, in this paper, we propose a novel Transformer-based Patch Refinement Model (\method) for the KGC task. The Transformer model is first proposed to handle Natural Language Processing (NLP) tasks~\cite{Transformer} and demonstrates superior capability in other visual tasks~\cite{imgtrans}. More recently, with the recent progress of Vision Transformer (ViT)~\cite{ViT}, attention-based modules achieve comparable or even better performances than their CNN counterparts on many vision tasks. Through attention mechanism, ViT-based models could dynamically focus on different embedding regions to obtain high-level informative features. What is more, ViT-based models do not impose any image-specific inductive bias, allowing them to handle a wider range of input data. Motivated by this, \method follows a ``Separate-and-Aggregate'' framework. In the separation stage, \method segments the input entity and relation embeddings into several patches. We explore three different separation schemes: \emph{1)} directly folding the embedding vector into several small patches;  \emph{2)} employing several trainable mapping matrices to obtain patches; and \emph{3)} using randomly initialized, but orthogonal mapping matrix to obtain patches. In the aggregation stage, unlike~\cite{ViT,clip} which use standard Transformer architecture, \method uses a cross-attentive architecture that deploys two separate attention modules to model the bi-directional interaction between the head entities and relations.

To evaluate our proposed approach, we conduct experiments on several benchmark datasets, including WN18RR, FB15k-237, YAGO37, and DB100K, for the KGC tasks. Our experiments show that 
\method successfully outperforms both non-Transformer-based and Transformer-based KGC methods, demonstrating the effectiveness of our approach. Our analysis shows the effectiveness of our cross-attention module design, patch-based position design, and embedding segmentation design. We find that \method is capable to learn useful KG knowledge using a large embedding dimension, while previous KGC models cannot. Finally, we demonstrate the advantages of \method in complex relation types, compared to previous KGC methods.

\section{Related Work}

\paragraph{Non-Neural-based methods.} A variety of non-neural based models are proposed for KGC leveraging simple vector space operations, such as dot product and matrix multiplication, to compute scoring function. TransE ~\cite{TransE} and its subsequent extensions~\cite{TransH,TransR} learn embeddings by representing relations as additive translations from head to tail entities. DistMult ~\cite{DistMult} uses multi-linear dot product to characterize three-way interactions among entities and relations. ComplEx ~\cite{ComplEx} represents entities and relations as complex-valued vectors, achieving an optimal balance between accuracy and efficiency. HolE ~\cite{HolE} utilizes cross-correlation, the inverse of circular convolution, for matching entity embeddings. More recently, SEEK~\cite{SEEK} proposes a framework for modeling segmented knowledge graph embeddings and demonstrates that several existing models, including DistMult, ComplEx, and HolE, can be considered special cases within this framework.

\paragraph{Neural-based methods.} Neural network (NN) based methods have also been explored. Approaches such as~\cite{dong2014knowledge,ravishankar2017revisiting} employ a Multi-Layer Perceptron (MLP) to model the scoring function. Moreover, Convolutional Neural Networks (CNN) have been utilized for KGC tasks. ConvE ~\cite{ConvE} aplies convolutional filters over reshaped head and relation embeddings to compute an output vector, which is then compared with all other entities in the knowledge graph. Subsequent works, including  ConvR~\cite{ConvR} and InteractE~\cite{InteractE} enhance ConvE by fostering interactions between head and relation embeddings.

\paragraph{Transformer-based methods.}


The Transformer model known for employing self-attention to process token sequences has achieved remarkable success in NLP tasks. This success is attributed not only to its capacity for handling long-range dependencies but also to its tokenization concept. Recently, this concept has been extended to other domains, such as computer vision through Vision Transformers~\cite{ViT} and multi-modality with Two-stream Transformers~\cite{vlbert}. These approaches have a common thread: they decompose the data (text or images) into smaller patches and process them using attention mechanisms. In the field of KGC, recent works have incorporated textual information and viewed entity and relation as the corresponding discrete descriptions. These methods often utilize pre-trained Transformers for encoding. However, high-quality textual KG data is not always accessible. As a result, our proposed method eschews additional textual information, instead integrating the tokenization concept into KGC to enhance performance.

\section{Method}
\subsection{Knowledge Graph Completion}
A \emph{Knowledge Graph} can be represented as ($\mathcal{E}$, $\mathcal{R}$, $\mathcal{T}$) where $\mathcal{E}$ and $\mathcal{R}$ denote the sets of entities and relations respectively. $\mathcal{T}$ is a collection of tuples $[(h, r, t)_i]$ where head and tail entity $h$, $t \in \mathcal{E}$ and relation $r \in \mathcal{R}$. The task of \textit{Knowledge Graph Completion} includes the head-to-tail prediction (e.g., predicting the head entity $h$ in the query $(?, r, t)$) and the tail-to-head prediction (e.g., predicting the tail entity $t$ in the query $(h, r, ?)$).

In this paper, following previous works~\cite{TransE,ConvE,ConvR}, we represent head and tail entities $h$ and $t$ as $\ve_h$ and $\ve_t \in \R^{d_e}$ and relation $r$ as $\ve_r \in \R^{d_r}$. Our objective is to learn a function $\mathrm{F}: \R^{d_e} \times \R^{d_r} \rightarrow \R^{d_e}$ such that given tuple $(h, r, t)$, the output of $\mathrm{F}(\ve_h, \ve_r)$ closely approximates $\ve_t$. For tail-to-head prediction, we additionally generate the reversed tuple $(t, r^{-1}, h)$ and train the output of $\mathrm{F}(\ve_t, \ve_{r^{-1}})$ to be closed to $\ve_h$.

\subsection{\method}
\begin{figure}[t]
\includegraphics[width=\textwidth]{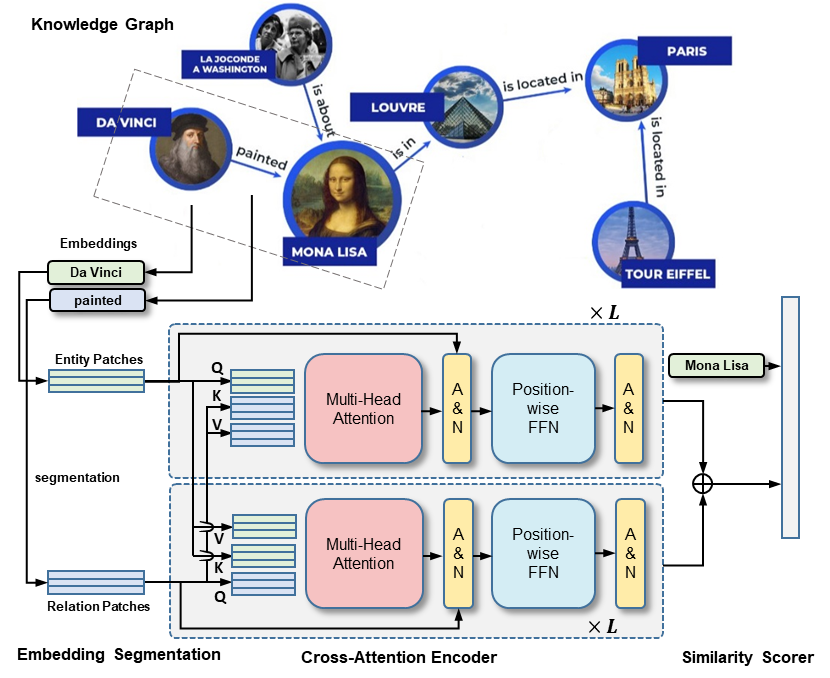}
\caption{An overview of \method} \label{fig:model_overview}
\end{figure}

In this section, we will introduce the details of \method. Fig.\ref{fig:model_overview} shows the overview of our \method model, which comprises three components: \textit{Embedding Segmentation}, \textit{Cross-Attention Encoder}, and \textit{Similarity Scorer}.

\subsubsection{Embedding Segmentation.} 
At this stage, \method converts entity and relation embeddings into sequences of patches. Formally, a segmentation function $pat(\cdot)$ is defined as follows:
\begin{align}
    \vp_{0}, \vp_{1}, \cdots, \vp_{k} = pat(\ve) 
\end{align}
where $\ve \in \mathbb{R}^{k\cdot d}$ is the input entity or relation embeddings. $\vp_i \in \mathbb{R}^{d}$ are segmented patches. $k$ is the sequence length of the generated patches and $d$ is the dimension of each patch.  Our method considers three segmentation variants, as shown in Fig.~\ref{fig:embed_seg}: 

\begin{figure}[t]
\includegraphics[width=\textwidth]{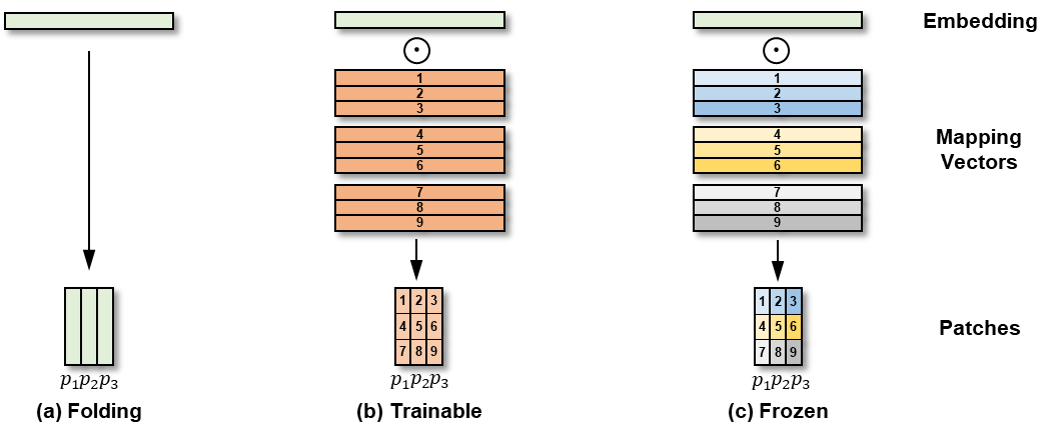}
\caption{Variants for Embedding Segmentation. $\odot$ denotes dot product operation. The mapping vectors with similar color (blue, yellow, grey) of frozen segmentation are mutually orthogonal. } \label{fig:embed_seg}
\end{figure}

\paragraph{Folding} involves reshaping the original embeddings $\ve$ into a sequence of equally-sized, smaller patches. Formally,
\begin{align}
pat(\cdot): \vp_{i, j} = \ve_{i * d + j}
\end{align}

\paragraph{Trainable Segmentation} employs a set of mapping vectors $v$ with adaptable parameters, enabling the model to learn and optimize the mapping function during training. This function can be written as: 
\begin{align}
pat(\cdot): \vp_{i, j} = u_{i, j} \odot \ve
\end{align}
where $u_{i, j}$ are trainable vectors.

\paragraph{Frozen Segmentation} utilizes the function with fixed parameters, precluding updates during the training process. Notably, the frozen Segmentation function comprises a set of matrices populated with mutually orthogonal vectors. This design choice aims to facilitate the generation of embedding patches that capture distinct aspects of an entity or relation, thereby enhancing the model's ability to represent diverse features. The patches are generated by:
\begin{align}
pat(\cdot): \vp_{i, j} = u_{i, j} \odot \ve \mathrm{\ ,where\ } u_{i, j} \odot u_{i, k} = 0 \mathrm{\ for\  all\ } j, k
\end{align}
The value of $u_{i, j}$ is obtained from the orthogonal matrix $U_{i}$, which is generated through singular value decomposition (SVD) of a randomly initialized matrix $M_i$. i.e.,
\begin{align}
    M_{i} = U_{i}\Sigma V_{i}^\top
\end{align}

\subsubsection{Cross-Attention Patch Encoder.} 
After segmenting entity and relation embedding into patches, we then aggregate these patches together via \emph{Cross-Attention Patch Encoder} which is based on a Siamese-Transformer architecture. We will discuss its details below.

\paragraph{Positional Embedding.}
The original Transformer model encodes them with either fixed or trainable positional encoding to preserve ordering information. However, unlike visual patches from images or words from the text, in the \method model, the patches from embeddings do not hold any much spatial information (i.e., the values in the first and last dimension alone do not carry particular semantic meaning). We thus remove the positional embedding in \method. We verify the effectiveness of this design in Section~\ref{sec:analysis}.

\paragraph{Cross-Attention Layer.}
Our proposed cross-attention layer process the entity and relation patches interactively with two separated attention modules:
\begin{equation}
\vh_{h}^i=\left\{
\begin{array}{lll}
\mathrm{MHA_{ER}^i}(\vh_{h}^{i-1}, \vh_{r}^{i-1}, \vh_{r}^{i-1}) & & {i > 0}  \\
pat(e_h) & & {i = 0} \\
\end{array} \right.
\end{equation}

\begin{equation}
\vh_{r}^i=\left\{
\begin{array}{lll}
\mathrm{MHA_{RE}^i}(\vh_{r}^{i-1}, \vh_{h}^{i-1}, \vh_{h}^{i-1}) & & {i > 0}  \\
pat(e_r) & & {i = 0} \\
\end{array} \right.
\end{equation}
where $\vh_{h}^{i}, \vh_{r}^{i}$ denote hidden representation of the $i$-th layer for head entity and relation respectively. $\mathrm{MHA_{ER}}$ and $\mathrm{MHA_{RE}}$ denotes Entity-to-Relation and Relation-to-Entity Attention module respectively. Both modules are based on the multi-head attention (MHA) mechanism, though they have different sets of parameters and inputs. The MHA module operates as follows:
\begin{align}
    \mathrm{MHA}(Q,K,V) = \mathrm{Concat(head_1, head_2, \cdots, head_H)}W^{o}, \\
    \mathrm{where\ head_i = Attention}(QW_i^{Q}, KW_i^{K}, VW_i^{V})  
\end{align}
$W_i^{Q} \in \mathcal{R}^{d\times d_{s}}$, $W_i^{K} \in \mathcal{R}^{d\times d_{s}}$, $W_i^{V} \in \mathcal{R}^{d\times d_{s}}$ are projection matrix. $d_{s} = d / H$ where $H$ is the predefined number of attention heads. $\mathrm{Attention}(\cdot)$ is the scaled dot-product attention module:
\begin{align}
    \mathrm{Attention}(Q, K, V) = \mathrm{softmax}(\frac{QK^\intercal}{\sqrt{d}})V \label{eq:attention}
\end{align}
where $Q \in \mathcal{R}^{N \times d}$, $K \in \mathcal{R}^{M \times d}$, $V \in \mathcal{R}^{M \times d}$, and $N$ and $M$ denote the lengths of queries and keys (or values).


\paragraph{Position-wise Feed-Forward Network Layer.}
The position-wise feed-forward network (FFN) refers to fully connected layers, which perform the same operation on each position of the input independently. 
\begin{align}
    \mathrm{FFN(X) = ReLU}(XW_1 + b_1)W_2 + b_2 
\end{align}
where $X$ is the output of the \emph{Cross-Attention Layer} i.e., $\vh_{h}^{i}$ or $\vh_{r}^{i}$. $W_1 \in \mathcal{R}^{d \times d_f}$, $b_1 \in \mathcal{R}^{d_f}$, $W_2 \in \mathcal{R}^{d_f \times d}$, $b_2 \in \mathcal{R}^{d}$ are trainable weights and bias. To facilitate the optimization on deep networks, \method employs a residual connection~\cite{residual} and Layer Normalization~\cite{layernorm} on \emph{Corss-Attention Layer} and \emph{FFN}. 


\subsubsection{Similarity Scorer.}
We employ a scoring function to evaluate the relevance between the output from the Cross-Attention Encoder and the target entity embedding. Specifically, we concatenate the hidden representations obtained from the two Transformers sub-modules and project them back to the entity dimension using a linear layer. 
\begin{align}
    \ve^{\prime} = \mathrm{Concat}(\overline{X_e}, \overline{X_r})W_o + b_o
\end{align}
In this context, $W_o \in \mathcal{R}^{(d_e + d_r) \times d_e}$, $b_o \in \mathcal{R}^{d_e}$ are weights and bias of the linear layer, respectively. $\overline{\cdot}$ is the operation to reshape Transformer output into a vector. Subsequently, we compute the dot product of the projected vector $\ve^{\prime}$ and the target entity embedding $\ve_{t}$. A sigmoid function is then applied to the result to ensure the final output falls within the $[0,1]$ range.This scorer can be expressed as:
\begin{align}
    s = \mathrm{Sigmoid}(\ve^{\prime} \odot \ve_{t})
\end{align}
 Algorithm \ref{algo:model} provides a full procedure of our proposed \method method.

\begin{algorithm}
\caption{\method for Computing the Score of a Triple in a KG}
\label{algo:model}
\scriptsize
    \begin{algorithmic}[1]
    \renewcommand{\algorithmicrequire}{\textbf{Input:}}
    \renewcommand{\algorithmicensure}{\textbf{Output:}}
    \REQUIRE Embedding for entities and relations, $E$ and $R$; head entity $h$, relation $r$ and tail entity $t$; tokenization function $tok(\cdot)$ 
    \ENSURE the score of triple $(h, r, t)$
    \STATE $\ve_h, \ve_r, \ve_t \gets E.get(h), R.get(r), E.get(t)$ \# get embeddings for h, r and t 
    \STATE $\ve_h \gets tok(\ve_h)$ 
    \STATE $\ve_r \gets tok(\ve_r)$ 
    \FOR{$i = 1\ to\ L$} 
         \STATE $\ve_h \gets \mathrm{LayerNorm}(\mathrm{MHA}(\ve_h, \ve_r, \ve_r) + \ve_h) $
         \STATE $\ve_h = \mathrm{LayerNorm}(\mathrm{FFN}(\ve_h) + \ve_h)$
         \STATE $\ve_r \gets \mathrm{LayerNorm}(\mathrm{MHA}(\ve_r, \ve_h, \ve_h) + \ve_r) $
         \STATE $\ve_r = \mathrm{LayerNorm}(\mathrm{FFN}(\ve_r) + \ve_r)$
    \ENDFOR
    \STATE $e^{\prime} \gets \mathrm{Concat}(\overline{e_h}, \overline{e_r})W_o + b_o$
    \STATE $s \gets \mathrm{Sigmoid}(e^{\prime} \odot \ve_{t})$
    \RETURN{$s$}
    \end{algorithmic}
\end{algorithm}
\subsection{Training and Inference}
For training, we leverage the standard binary cross entropy loss with label smoothing:
\begin{align}
\mathcal{L}_{\text{BCE}} = -\frac{1}{N}\sum_{i=1}^N [y_i \log(s_i) + (1-y_i) \log(1-s_i)]
\end{align}
where $p_i$ and $y_i$ are the score and label of the \textit{i}-th training instance respectively. $y_i \in [\epsilon, 1-\epsilon]$, where $\epsilon$ is the label smoothing value. For inference, \method computes the scores of the query $(h, r, ?)$ for every possible entities and rank them based on the corresponding scores. More details are presented in Section \ref{sec:Experimental Setup}.

\section{Experimental Results}
In this section, we evaluate \method against various baselines in the KGC task on multiple benchmark KGs. 
\subsection{Experimental Setup}\label{sec:Experimental Setup}

\paragraph{Dataset.} Our proposed method is evaluated on four publicly available benchmark datasets: \datafbn{}~\cite{FB15K237}, \datawnn{}~\cite{ConvE}, \datayg{}~\cite{DB100K} and \datadb{}~\cite{YAGO37}. A summary of these datasets is provided in Table \ref{tab:datasets}. \datafbn{} and \datawnn{} are widely-used benchmarks derived from FB15K and WN18~\cite{TransE}, respectively. They are free from the inverting triples issue. \datafbn{} and \datawnn{} were created by removing the inverse relations from FB15K and WN18 to address this issue. \datadb{} and \datayg{} are two large-scale datasets. \datadb{} was generated from the mapping-based objects of core DBpedia~\cite{DBPedia}, while \datayg{} was extracted from the core facts of YAGO3~\cite{YAGO3}. 

\begin{table*}[!ht]
\vspace{-1.0em}
\centering
\caption{Statistics of datasets.}\label{tab:datasets}
\begin{tabular}{ p{5.2em} | p{1.4cm}<{\raggedleft} p{1.0cm}<{\raggedleft}p{1.4cm}<{\raggedleft}p{1.4cm}<{\raggedleft}p{1.4cm}<{\raggedleft}}
\toprule
\textbf{Dataset} & \textbf{\#Ent} & \textbf{\#Rel} & \textbf{\#Train} & \textbf{\#Valid} & \textbf{\#Test} \\
\midrule
WN18RR & $40,943$ & $11$ & $86,835$ & $3,034$ & $3,134$ \\
FB15K-237 & $14,541$ & $237$ & $272,115$ & $17,535$ & $20,466$  \\
DB100K & $99,604$ &  $470$ &  $597,572$ &  $50,000$ &  $50,000$ \\
YAGO37 & $123,189$ & $37$ & $989,132$ & $50,000$ & $50,000$ \\
\bottomrule
\end{tabular}
\vspace{-1.0em}
\end{table*}


\paragraph{Evaluation protocol.} Our experiment follows the filtered setting proposed in~\cite{TransE}. Specifically, for each test triple $(h, r, t)$, two types of triple corruption are considered, i.e., tail corruption $(h, r, ?)$ and $(t, r^{-1}, ?)$. Every possible candidate in the knowledge graph is used to replace the entity, forming a set of valid and invalid triples. The goal is to rank the test triple among all the corrupted triples. In the filtered setting, any true triples observed in the train/validation/test set except the test triple $(h, r, t)$ are excluded during evaluation. The evaluation metrics include the mean reciprocal rank (MRR) and the proportion of correct entities ranked in the top $n$ (H@n) for $n = 1, 3, 10$. The evaluation is performed over all test triples on both types of triple corruption.




\begin{table*}[!ht]
\vspace{-1.0em}
\centering
\caption{Optimal hyperparameters for various KGC benckmarks}\label{tab:optimal_param}
\begin{tabular}{ p{5.2em} | p{3.2em}<{\centering}p{3.2em}<{\centering}p{3.2em}<{\centering}p{3.2em}<{\centering}p{3.2em}<{\centering}p{3.2em}<{\centering}p{3.2em}<{\centering} }
\toprule
& $\eta$ &$L$ &$d_e$ &$d_r$ &$p_1$ &$p_2$ &$p_3$ \\
\midrule
WN18RR &1e-3 &2 &100 &5000 &0.1 &0.1 &0.4  \\ 
FB15K-237 &1e-3 &12 &100 &3000 &0.3 &0.1 &0.4   \\ 
DB100K&5e-4 &4 &200 &5000 &0.1 &0.1 &0.4  \\ 
YAGO37&1e-4 &4 &200 &1000 &0.1 &0.1 &0.1  \\ 
\bottomrule
\end{tabular}
\vspace{-1.0em}
\end{table*}
\paragraph{Implementation details.} 
We implement \method in PyTorch\footnote{\url{https://pytorch.org/}}. In this experiment, we fix mini-batch size $\mathcal{B}$ to 256, Transformer dimensions $d$ to 50, and label smoothing value $\epsilon$ to 0.1. The other hyper-parameters are tuned via grid search. Specifically, we select learning rate $\eta$ from $\{$1e-4, 5e-4, 1e-3$\}$, number of layers $L$ from $\{2, 4, 12\}$, entity embedding size $d_e$ $\{100, 200\}$, relation embedding size $d_r$ $\{1000, 3000, 5000\}$. All dropout ratios, i.e., $p_1$ on embedding patches, $p_2$ on Cross-Attention Encoder and $p_3$ on the linear layer in Similarity Scorer, are tuned in $\{0.1, 0.2, 0.3, 0.4\}$. We use Adam ~\cite{Adam} to optimize our model. On each dataset, we select the optimal configuration according to the best MRR on the validation set within $2500$ epochs.  The optimal configurations of \method on the four datasets are listed in Table \ref{tab:optimal_param}.

\subsection{Experimental Results}
\begin{table*}[t]
\vspace{-1.0em}
\centering
\caption{Experimental results of baseline models on \datafbn{}, \datawnn{}.}
\label{tab:main_result_wnfb} 
\resizebox{\columnwidth}{!}{
\begin{tabular}{lp{3.2em}<{\centering}p{3.2em}<{\centering}p{3.2em}<{\centering}p{3.2em}<{\centering} p{3.2em}<{\centering}p{3.2em}<{\centering}p{3.2em}<{\centering}p{3.2em}<{\centering}p{0.1em}<{\centering} }
\toprule
& \multicolumn{4}{c}{\textbf{\datafbn{}}} & \multicolumn{4}{c}{\textbf{\datawnn{}}} &  \\ 
\cmidrule(r){2-5}  \cmidrule(r){6-9} 
&MRR &H@1 &H@3 &H@10 &MRR &H@1 &H@3 &H@10  \\
\midrule
\textbf{\emph{\footnotesize{Non-Transformer-Based Methods}}}\\
TransE \cite{TransE}    &.279 &.198 &.376 &.441 &.243 &.043 &.441 &.532 \\
DistMult \cite{DistMult} 	&.241 &.155 &.263 &.419 &.430 &.390 &.440 &.490 \\
ComplEx	\cite{ComplEx}	&.247 &.158 &.275 &.428 &.440 &.410 &.460 &.510 \\
R-GCN \cite{R-GCN} 		&.249 &.151 &.264 &.417 &- &- &- &- \\
SACN \cite{SACN} 		&.350 &.260 &.390 &.540   &.470 &.430  &.480 &.540 \\
ConvR \cite{ConvR}		&.350 &.261 &.385 &.528  &.475 &.443 &.489 &.537 \\
RotatE \cite{RotatE}    &.338 &.241 &.375 &.533 &.476 &.428 &.492 &.571 \\
ConvE \cite{ConvE} 		&.325 &.237 &.356 &.501 &.430 &.400 &.440 &.520 \\
InteractE \cite{InteractE}&.354 &.263 &- &.535 &.463 &.430 &- &.528 \\
AcrE \cite{AcrE} &.358 &.266 &.393 &.545 &.459 &.422 &.473 &.532 \\
\midrule 
\textbf{\emph{\footnotesize{Transformer-based methods}}} \\
KG-BERT~\cite{KG-BERT} &- &- &- &.420 &.216 &.041 &.302 &.524 \\
MTL-KGC~\cite{MTL-KGC} &.267 &.172 &.298 &.458 &.331 &.203 &.383 &.597 \\
StAR~\cite{StAR} &.296  &.205 &.322 &.482 &.401 &.243 &.491 &\textbf{.709} \\
GenKGC~\cite{GenKGC} &- &.192 &.355 &.439 &- &.287 &.403 &.535  \\
\midrule 
\method{} (ours) &\textbf{.364} &\textbf{.271} &\textbf{.400}  &\textbf{.551} &\textbf{.480} &.439 &\textbf{.499} &.558 \\ 
\bottomrule
\end{tabular}}
\vspace{-1.0em}
\end{table*}
Table \ref{tab:main_result_wnfb} presents a comprehensive comparison of our proposed \method model, against the baseline models on two popular \datafbn{} and \datawnn{} benchmarks. Our experimental results indicate that \method is highly competitive against the state-of-the-art models. Specifically, \method achieves improvements of 0.009 in MRR and 0.6\% in H@10 compared to the previous models used on \datawnn{}. On \datawnn{}, \method obtains better results in terms of MRR (0.480 vs. 0.476) and H@3 (0.499 vs. 0.492) and is competitive in the H@10 and H@1 metrics. We attribute this discrepancy to the fact that the WN18RR dataset is a lexicon knowledge graph that relies heavily on textual information. 
As a result, the KGC models that incorporate pre-trained language models, such as StAR and MTL-KGC, achieve better performance than \method in those metrics.

\begin{table*}[t]
\vspace{-1.0em}
\centering
\caption{Experimental results of several models evaluated on \datadb, \datayg{}. }\label{tab:main_result_dbyg} 
\resizebox{\columnwidth}{!}{
\begin{tabular}{p{11em}p{3.2em}<{\centering}p{3.2em}<{\centering}p{3.2em}<{\centering}p{3.2em}<{\centering}p{3.2em}<{\centering}p{3.2em}<{\centering}p{3.2em}<{\centering}p{3.2em}<{\centering}p{0.1em}<{\centering}}
\toprule
& \multicolumn{4}{c}{\textbf{\datadb{}}} & \multicolumn{4}{c}{\textbf{\datayg{}}} &  \\ 
\cmidrule(r){2-5}  \cmidrule(r){6-9} 
&MRR &H@1 &H@3 &H@10 &MRR &H@1 &H@3 &H@10  \\
\midrule
TransE \cite{TransE} &.111 &.016 &.164 &.270 &.303 &.218 &.336 &.475 \\
DistMult \cite{DistMult} &.233 &.115 &.301 &.448 &.365 &.262 &.411 &.575 \\
HolE \cite{HolE} &.260  &.182 &.309 &.411 &.380 &.288 &.420 &.551 \\
ComplEx \cite{ComplEx} &.242 &.126 &.312 &.440 &.417 &.320 &.471 &.603 \\
Analogy \cite{ANALOGY} &.252 &.142 &.323 &.427 &.387 &.302 &.426 &.556 \\
SEEK \cite{SEEK} &.338 &.268 &.370 &.467 &.454 &.370 &.498 &.622 \\
AcrE \cite{AcrE} &.413 &.314 &.472 &.588 &- &- &- &- \\
\midrule 
\method{} (ours)	&\textbf{.436} &\textbf{.353} &\textbf{.479}  &\textbf{.589} &\textbf{.523} &\textbf{.449} &\textbf{.567} &\textbf{.656}  \\
\bottomrule
\end{tabular}}
\vspace{-1.0em}
\end{table*}
To further verify the effectiveness of \method on larger KG, we evaluate our method on \datadb{} and \datayg{}. Table \ref{tab:main_result_dbyg} presents the performance comparison of \method with other baseline KGC models. On both benchmarks, \method outperforms existing methods on all evaluation criteria.
In particular, \method demonstrates superiority on \datayg{} with a significant relative improvement of 15.2\% (0.523 vs 0.454) and 5.5\% (0.656 vs 0.622) in MRR and H@10 respectively. These findings indicate the feasibility and applicability of \method on real-world large-scale knowledge graphs.

\section{Analysis}\label{sec:analysis}
In this section, we investigate \method from various perspectives. In the first place, we show the effectiveness of the design choices in \method. We then show that \method is capable of capturing more knowledge via a large embedding dimension. Finally, we demonstrate the advantages of \method in complex knowledge relations. All experiments are conducted on \datafbn{}.

\subsection{Impact of Cross Attention} 
In this section, we aim to examine the effectiveness of cross-attention in our proposed model by comparing it with two variants: 1) full self-attention, in which entity and relation patches are combined together before being fed into the model, and full self-attention is applied on the combined input; and 2) separate self-attention, in which each Transformer conducts self-attention on entity and relation patches independently before concatenating their results in the Similarity Scorer. The experimental results demonstrate that our proposed cross-attention method outperforms both the full self-attention and separate self-attention variants. We hypothesize that the cross-attention mechanism only learns to connect patches from different embeddings (i.e., patches from the same embedding never interact with each other), avoiding unnecessary interference from a single embedding. This could be the primary reason why cross-attention outperforms the full self-attention variant. Furthermore, the separate self-attention variant lacks interaction between entities and relations, which could explain the significant performance drop. 
\begin{table*}[!ht]
\vspace{-1.0em}
\centering
\caption{Analysis for model structure on \datafbn{}. Att. denotes attention.}\label{tab:analysis_structure}
\begin{tabular}{ p{7em} | p{1em}p{5.2em}p{5.2em}p{5.2em}p{5.2em} }
\toprule
&& \textbf{MRR} & \textbf{H@1} & \textbf{H@3} & \textbf{H@10} \\
\midrule
\method &&$.3640$ &$.2708$ &$.3997$ &$.5506$ \\
Full Self-Att. &&$.3599_{\downarrow .0041}$ &$.2656_{\downarrow .0052}$ &$.3966_{\downarrow .0031}$ &$.5476_{\downarrow .0030}$  \\ 
Sep. Self-Att. &&$.3387_{\downarrow .0253}$ &$.2503_{\downarrow .0205}$ &$.3698_{\downarrow .0299}$ &$.5161_{\downarrow .0345}$ \\ 
\bottomrule
\end{tabular}
\vspace{-1.0em}
\end{table*}



\subsection{Impact of Positional Encoding} 
The original Transformer model~\cite{Transformer} involves positional encoding to convey positional information of sequential tokens. To examine the impact of positional encoding on \method, we conduct an experiment with two variants: 1) trainable positional encoding (TPE) and 2) fixed positional encoding (FPE). Our experimental results demonstrate that the model without positional encoding (\method) outperforms the other two variants. We believe that this is due to the nature of embeddings patches, which inherently capture the features of entities or relations in a non-sequential manner. As a result, integrating  positional encoding into the model introduces extraneous  positional information, causing a decline in performance.
\begin{table*}[!ht]
\vspace{-1.0em}
\centering
\caption{Analysis for positional encoding (PE) on \datafbn{}. Our proposed \method does not apply PE.}\label{tab:analysis_pe}
\begin{tabular}{ p{6.7em} | p{1em}p{5.2em}p{5.2em}p{5.2em}p{5.2em}}
\toprule
Models && \textbf{MRR} & \textbf{H@1} & \textbf{H@3} & \textbf{H@10} \\
\midrule
\method &&.3640 &.2708 &.3997 &.5506 \\
w/ TPE &&$.3354_{\downarrow .0286}$ &$.2474_{\downarrow .0234}$ &$.3660_{\downarrow .0337}$ &$.5107_{\downarrow .0399}$ \\ 
w/ FPE &&$.2580_{\downarrow .1060}$ &$.1897_{\downarrow .0811}$ &$.2789_{\downarrow .1208}$ &$.3907_{\downarrow .1599}$ \\ 
\bottomrule
\end{tabular}
\vspace{-1.0em}
\end{table*}

\subsection{Impact of Segmentation} 
In this section, we explore the impact of segmentation on our proposed model, specifically examining the performance without using segmentation, and employing folding, trainable, and frozen segmentation. Our experimental results in Table \ref{tab:analysis_seg} present that the utilization of segmentation yields a substantial performance improvement. With respect to the segmentation methods, frozen segmentation outperforms the other two variants. We believe this is due to the orthogonal vectors employed in frozen segmentation, which enhance the model's capacity to discern features of embeddings from distinct perspectives. Conversely, trainable segmentation, which allows parameters freely update during training, may face difficulties in achieving this. These findings emphasize the importance of selecting segmentation variants in the context of knowledge graph completion tasks. The superior performance of frozen segmentation suggests that these orthogonal vectors can be advantageous in extracting  diverse features from entity and relation embeddings.
\begin{table*}[!ht]
\vspace{-1.0em}
	\centering
 	\caption{Analysis for tokenization variants on \datafbn{}.}
	\label{tab:analysis_seg}
	\begin{tabular}{ p{6.7em} | p{1em}p{5.2em}p{5.2em}p{5.2em}p{5.2em}}
		\toprule
		 && \textbf{MRR} & \textbf{H@1} & \textbf{H@3} & \textbf{H@10} \\
		\midrule
        \method &&$.3640$ &$.2708$ &$.3997$ &$.5506$ \\
		w/o Seg. &&$.3501_{\downarrow .0139}$ &$.2592_{\downarrow .0116}$ &$.3850_{\downarrow .0147}$ &$.5316_{\downarrow .0190}$ \\ 
        Folding Seg. &&$.3623_{\downarrow .0017}$ &$.2695_{\downarrow .0013}$ &$.3979_{\downarrow .0018}$ &$.5488_{\downarrow .0018}$ \\ 
        Trainable Seg. &&$.3572_{\downarrow .0068}$ &$.2642_{\downarrow .0066}$ &$.3936_{\downarrow .0061}$ &$.5433_{\downarrow .0073}$ \\ 

		\bottomrule
	\end{tabular}
 \vspace{-1.0em}
\end{table*}

\subsection{Effectiveness of \method via a Large Relation Embedding Dimension} 
\begin{figure}
    \centering
    \begin{subfigure}[b]{0.48\textwidth}
        \includegraphics[width=\linewidth]{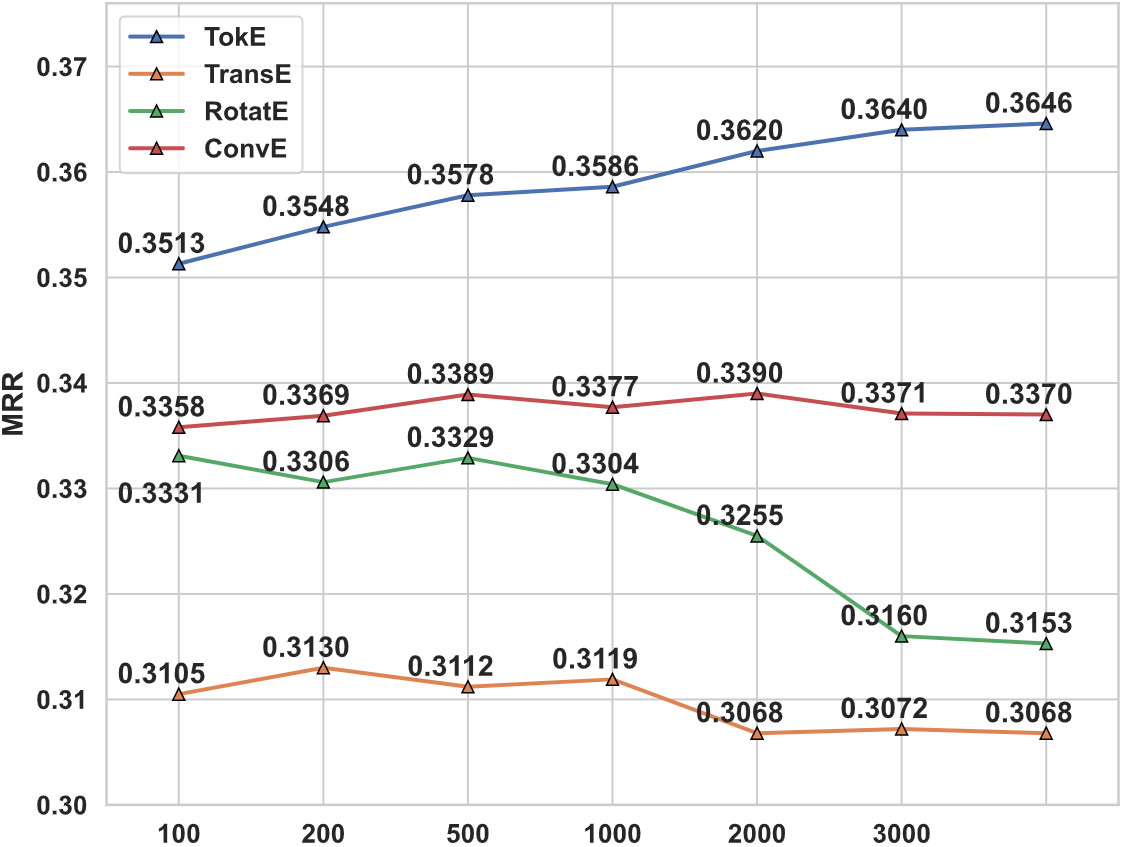}
        \label{fig:subfig1}
    \end{subfigure}
    \begin{subfigure}[b]{0.48\textwidth}
        \includegraphics[width=\linewidth]{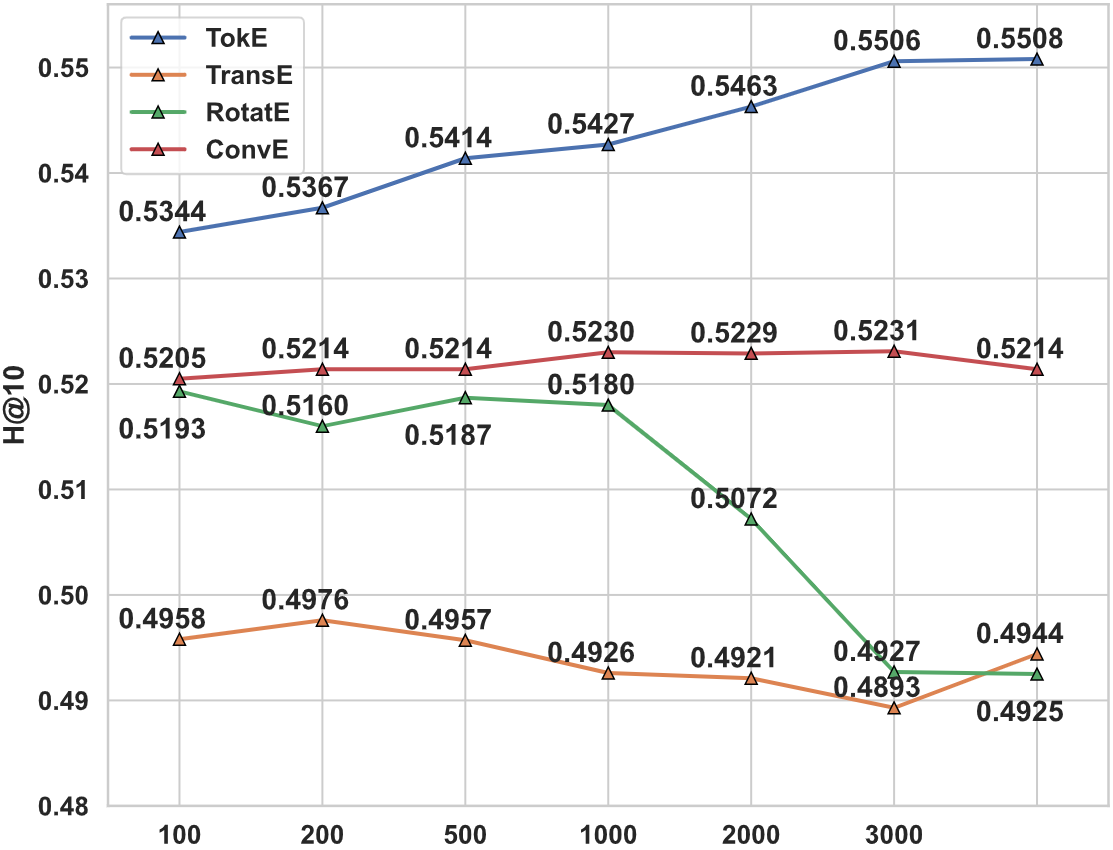}
        \label{fig:subfig2}
    \end{subfigure}
    \caption{Analysis of relation embedding size on \datafbn{}}
    \label{fig:relation embedding size}
\end{figure}

In a typical KG, the number of relations is much less than the number of entities. Thus, we hypothesize that the KGC models that can effectively handle a large relation embedding dimension should achieve superior KGC performance. We verify this hypothesis in this section. Fig.~\ref{fig:relation embedding size} shows a clear performance increasing trend for \method as the length of relation embeddings increases. However, the other baseline KGC models, such as TransE, ConvE, and RotatE, do not deliver similar improvement; RotatE even suffers from performance delegations
after the embedding dimension increases. This result shows that \method could capture more knowledge by using a large embedding dimension, while other methods cannot due to their insufficient modeling expressiveness. Such ability allows \method to capture more knowledge for relation embeddings and achieve better performance.


\subsection{Analysis on Different Types of Relations} 
In this section, we analyze the performance of different types of relations for various models: TransE, ConvE, RotatE and \method. To categorize the relations, we considered the average number of tails per head and heads per tail, grouping them into four distinct types: 1-1 (one-to-one), 1-N (one-to-many), N-1 (many-to-one), and N-N (many-to-many). The results presented in Table \ref{tab:analysis_nton} demonstrate that our \method model outperforms the other models in handling more complex relation types, such as 1-N, N-1, and N-N. This indicates that the increased interaction in our model allows it to capture intricate relationships more effectively. We note that TransE and ConvE perform better for simpler one-to-one relations. We believe there could be two reasons behind this phenomenon: 1) TransE and ConvE are intrinsically adept at representing simple relations (i.e., one-to-one), and 2) the limited number of evaluation instances for this category might result in biased results. Despite this, this experiment verifies the strength of our proposed \method model in modeling complex relation types and highlights its potential applicability to a wide range of more complicated KGC tasks.
\begin{table}
\vspace{-1.0em}
\centering
\caption{Experimental results by relation categories for KGC methods on \datafbn{}.}\label{tab:analysis_nton} 
\begin{tabular}{p{2.2em}p{3.2em}<{\centering}p{3.2em}<{\centering}p{3.2em}<{\centering}p{3.2em}<{\centering}p{3.2em}<{\centering}p{3.2em}<{\centering}p{3.2em}<{\centering}p{3.2em}<{\centering}p{3.2em}<{\centering}p{0.1em}<{\centering}}
\toprule
&& \multicolumn{2}{c}{\textbf{TransE}} & \multicolumn{2}{c}{\textbf{ConvE}} &
\multicolumn{2}{c}{\textbf{RotatE}} &
\multicolumn{2}{c}{\method} \\ 
\cmidrule(r){3-4} \cmidrule(r){5-6} \cmidrule(r){7-8} \cmidrule(r){9-10} 
&\#triples&MRR &H@10 &MRR &H@10  &MRR &H@10 &MRR &H@10\\
\midrule
1-1  &192 &\textbf{.4708} &.5520 &.4384 &.5546 &.3315 &.5078 &.3339 &\textbf{.5625} \\ 
1-N  &1,293 &.2388 &.3650 &.2532 &.3789 &.2719 &.4017 &\textbf{.2828} &\textbf{.4203} \\
N-1  &4,185 &.3975 &.4972 &.4151 &.5187 &.4207 &.5168 &\textbf{.4647} &\textbf{.5698} \\
N-N  &14,796 &.2877 &.5063 &.3133 &.5315 &.3167 &.5337 &\textbf{.3432} &\textbf{.5564} \\
\bottomrule
\end{tabular}
\vspace{-1.0em}
\end{table}
\section{Conclusion}
In this paper, motivated by the recent advances in Transformers, we propose a novel Transformer-based Patch Refinement model \method for knowledge graph completion. 
\method includes three main components: Embedding Segmentation, Cross-Attention Encoder, and Similarity Scorer. We first segment the knowledge graph embeddings into patches and then apply a Transformer-based cross-attention encoder to model interaction between entities and relations. Finally, the Similarity Scorer combines the encoded representations to compute the similarity between inputs and target entities.
The experiments on four benchmark datasets (WN18RR, FB15k-237, DB100K and YAGO37) show that our proposed \method outperforms existing state-of-the-art knowledge graph completion (KGC) approaches. These results validate the effectiveness of our approach and highlight the potential advantages of incorporating patch-based embeddings and cross-attention mechanisms in such tasks. 

\section{Acknowledgement}
This research is supported by the National Research Foundation, Singapore and Infocomm Media Development Authority under its Trust Tech Funding Initiative and Strategic Capability Research Centres Funding Initiative. Any opinions, findings and conclusions or recommendations expressed in this material are those of the author(s) and do not reflect the views of National Research Foundation, Singapore and Infocomm Media Development Authority.

\end{document}